\documentclass[conference]{IEEEtran}
\usepackage{times}

\usepackage[numbers]{natbib}
\usepackage{multicol}
\usepackage{epsfig} 
\usepackage{mathptmx} 
\usepackage{times} 
\usepackage{amsmath,bm} 
\usepackage{amssymb}  
\usepackage{dirtytalk}
\usepackage[makeroom]{cancel}
\usepackage{soul}
\usepackage{epigraph}
\usepackage{stmaryrd}
\usepackage{listings}

\usepackage{graphicx}
\usepackage{caption}
\usepackage{tabularx}
\usepackage{tikz}

\usepackage{times}
\usepackage[utf8]{inputenc} 
\usepackage[T1]{fontenc}    
\usepackage{url}            
\usepackage{booktabs}       
\usepackage{nicefrac}       
\usepackage{microtype}      
\usepackage{adjustbox}      
\usepackage{graphics,color}
\let\labelindent\relax 
\usepackage[shortlabels]{enumitem}
\usepackage{bbm}

\usepackage{xcolor}

\definecolor{codegreen}{rgb}{0,0.6,0}
\definecolor{codegray}{rgb}{0.5,0.5,0.5}
\definecolor{codepurple}{rgb}{0.58,0,0.82}
\definecolor{backcolour}{rgb}{0.95,0.95,0.92}

\lstdefinestyle{mystyle}{
    backgroundcolor=\color{backcolour},   
    commentstyle=\color{codegreen},
    keywordstyle=\color{magenta},
    numberstyle=\tiny\color{codegray},
    stringstyle=\color{codepurple},
    basicstyle=\ttfamily\footnotesize,
    breakatwhitespace=false,         
    breaklines=true,                 
    captionpos=b,                    
    keepspaces=true,                 
    numbers=left,                    
    numbersep=5pt,                  
    showspaces=false,                
    showstringspaces=false,
    showtabs=false,                  
    tabsize=2
}

\newcommand{\trsp}{{\scriptscriptstyle\top}}

\DeclareMathOperator*{\argmin}{arg\,min}

\newcommand{\surface}{\mathcal{S}}
\newcommand{\pose}{\bm T}

\newcommand{\vertexset}{V}
\newcommand{\vertexnum}[1][]{{|V_{#1}|}}
\newcommand{\vertexidx}{i}
\newcommand{\vertex}{\bm{v}}

\newcommand{\edgeset}[1][]{E_{#1}}
\newcommand{\edgenum}[1][]{|E_{#1}|}
\newcommand{\edgeidx}{j}
\newcommand{\edge}{\bm{e}}

\newcommand{\astar}{\bm{\alpha}^{\star}}
\newcommand{\astaru}{\bm{\alpha}^{\mathrm{u}\star}}
\newcommand{\astarc}{\bm{\alpha}^{\mathrm{c}\star}}
\newcommand{\astarucomp}{\alpha^{\mathrm{u}\star}}

\newcommand{\facenum}{|F|}

\newcommand{\primitiveset}{L}
\newcommand{\primitivenum}{|L|}
\newcommand{\primitiveidx}{l}

\newcommand{\contactpointnum}{|K|}
\newcommand{\contactpointidx}{k}

\newcommand{\polyhedronnum}{N}
\newcommand{\polyhedronidx}{i}

\newcommand{\opcnum}{N}
\newcommand{\opcidx}{i}

\newcommand{\binaryidx}{b}

\newcommand{\normal}{\bm{n}}
\newcommand{\point}{\bm{p}}
\newcommand{\signeddistance}{d}

\newcommand{\softness}{\tau}

\usepackage{xcolor}
\definecolor{ourblue}{rgb}{0.368,0.507,0.71}    
\definecolor{ourorange}{rgb}{0.881,0.611,0.142} 
\definecolor{ourgreen}{rgb}{0.56,0.692,0.195}   
\definecolor{ourred}{rgb}{0.923,0.386,0.209}    
\definecolor{ourviolet}{rgb}{0.528,0.471,0.701} 
\definecolor{ourbrown}{rgb}{0.772,0.432,0.102}  
\definecolor{ourazure}{rgb}{0.364,0.619,0.782}  
\definecolor{ourolive}{rgb}{0.572,0.586,0.}     
\definecolor{ourgray}{RGB}{102,88,84}           

\definecolor{ourblue2}{RGB}{9,134,223} 
\definecolor{ourdarkblue2}{RGB}{5,97,164} 
\definecolor{ourlightblue2}{RGB}{132,201,250} 

\definecolor{ourorange2}{RGB}{224,90,18} 
\definecolor{ourdarkorange2}{RGB}{160,63,9} 
\definecolor{ourlightorange2}{RGB}{246,175,137} 

\definecolor{ouryellow2}{RGB}{227,213,25} 
\definecolor{ourdarkyellow2}{RGB}{177,166,17} 
\definecolor{ourlightyellow2}{RGB}{242,235,140} 

\definecolor{ourpink2}{RGB}{247,24,139} 
\definecolor{ourdarkpink2}{RGB}{164,4,86} 
\definecolor{ourlightpink2}{RGB}{250,163,207} 

\definecolor{ourgreen2}{RGB}{159,198,52} 
\definecolor{ourdarkgreen2}{RGB}{109,138,30} 
\definecolor{ourlightgreen2}{RGB}{209,228,154} 

\definecolor{ourgray2}{RGB}{124,124,115} 
\definecolor{ourdarkgray2}{RGB}{87,87,81} 
\definecolor{ourlightgray2}{RGB}{194,194,189} 

\usepackage[pagebackref,breaklinks,colorlinks,allcolors=ourblue,bookmarks=true]{hyperref}
\usepackage[capitalise]{cleveref}

\begin{document}

\title{Smoothly Differentiable and Efficiently Vectorizable Contact Manifold Generation}

\author{\normalsize
        Onur Beker$^{1}$,
        A.\ Ren\'e Geist$^{1}$,
        Anselm Paulus$^{1}$,
        Nico Gürtler$^{1}$,
        Ji Shi$^{1}$,
        Sylvain Calinon$^{2,3}$,
        Georg Martius$^{1}$ \\
        $^{1}$University of Tübingen \quad 
        $^{2}$Idiap Research Institute \quad 
        $^{3}$EPFL
        }
        
\maketitle

\begin{abstract}
Simulating rigid-body dynamics with contact in a fast, massively vectorizable, and smoothly differentiable manner is highly desirable in robotics. An important bottleneck faced by existing differentiable simulation frameworks is contact manifold generation: representing the volume of intersection between two colliding geometries via a discrete set of properly distributed contact points. A major factor contributing to this bottleneck is that the related routines of commonly used robotics simulators were not designed with vectorization and differentiability as a primary concern, and thus rely on logic and control flow that hinder these goals. We instead propose a framework designed from the ground up with these goals in mind, by trying to strike a middle ground between: i) convex primitive based approaches used by common robotics simulators (efficient but not differentiable), and ii) mollified vertex-face and edge-edge unsigned distance-based approaches used by barrier methods (differentiable but inefficient). Concretely, we propose: i) a representative set of smooth analytical signed distance primitives to implement vertex-face collisions, and ii) a novel differentiable edge-edge collision routine that can provide signed distances and signed contact normals. The proposed framework is evaluated via a set of didactic experiments and benchmarked against the collision detection routine of the well-established Mujoco XLA framework, where we observe a significant speedup.  
Supplementary videos can be found at \href{https://github.com/bekeronur/contax}{https://github.com/bekeronur/contax}, where a reference implementation in JAX will also be made available at the conclusion of the review process.
\end{abstract}

\IEEEpeerreviewmaketitle

\section{Introduction}
\begin{figure*}[th]
\centering
\includegraphics[width=\textwidth]{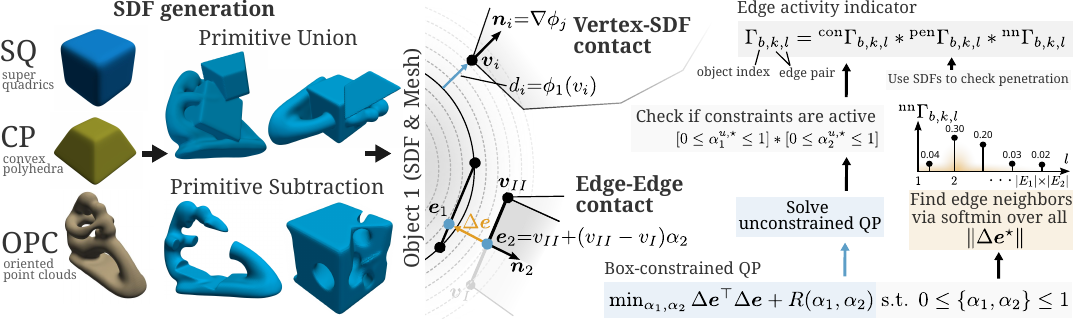}
\caption{We introduce a contact manifold generation framework that is smoothly differentiable, fast, and well-suited for vectorization. This is enabled by two contributions. The first is a pipeline for constructing signed distance fields (SDF) for arbitrary surfaces from a representative set of analytical primitives. These SDFs then serve two roles: (i)~acting as a differentiable acceleration structure to efficiently select vertex–face (V–F) and edge–edge (E–E) contact candidates, and (ii)~providing signed distances, normals, and activity indicators for V–F and E–E interactions in a way that retains curvature information (since by default, they happen between flat, planar facets). The second contribution is a method to make witness point computations for V–F and E–E pairs smoothly differentiable by solving the associated quadratic programs via an analytical active-set approach.}
\vspace{-5mm}
\label{fig:main}
\end{figure*}

What characteristics define a good physics simulator for robotics? It is appropriate to split such desiderata into two groups based on their underlying motivation: \say{physics for prediction}, and \say{physics for control}. When simulating physics for prediction, there already exists an intelligent specification (e.g., a design or a behavior) driving the simulation, and one mainly wants to observe an approximation of its consequences in a way that achieves a notion and degree of realism dictated by the downstream application. For example, in the context of physically-based animation \cite{baraff1997, andrews2022contact}, it is the animator who sets up a scene or a character and specifies its behavior. In the context of computer-aided engineering software \cite{mscadams}, it is the responsibility of a mechanical engineer to specify an intelligent design to be tested. In the context of physics engines for computer games \cite{gregory2018game}, the burden of intelligent conduct falls on the player. When simulating physics for control, however, the desired consequences are determined first, and one instead wants to infer an intelligent specification (e.g., system parameters to be identified or actuator control signals to be commanded) that will drive the simulation to incite said consequences.\looseness-1

While physics for prediction purposes should be formulated with realism as the primary goal, formulations of physics for control purposes should instead actively try to identify the right simplifications to reality in order to reduce the complexity of searching for the desired intelligent specification. Particularly within the context of differentiable rigid body dynamics simulation with contact, this involves making the appropriate relaxations to the discontinuous reality of contact interactions in order to create a well-behaved optimization landscape. But what are these appropriate relaxations, and where should they happen in the overall simulation process? 

The numerous computations that make up any typical rigid-body dynamics simulation routine can be grouped into two main subroutines \cite{baraff1997, andrews2022contact}: i)~\emph{collision detection} to obtain geometric measurements about relative configurations of the simulated surfaces that is useful for transcribing contact constraints, and ii)~\emph{contact dynamics} to timestep the simulation in a way that respects said contact constraints. Naturally, relaxations have to be made on both of these fronts to ensure the differentiability of the whole simulation process.  This paper focuses on the collision detection part of the overall problem.

Looking at the collision detection routines of commonly used simulation frameworks in robotics \cite{todorov2012mujoco, NVIDIA_Isaac_Sim, coumans2016pybullet, drake}, one can observe that most of the underlying algorithms \cite{sutherland1974reentrant, gilbert1988fast, van2001proximity, gregorius2015robust} were conceived from a physics for prediction perspective (often within a computer graphics, video game design, or computational science and engineering context), and are optimized for speed and minimal memory footprint. This creates complications when they are employed for generating intelligent robot behavior. For example, such algorithms often involve significant branching and control flow, which result in pathological search landscapes with many discontinuities that throw off numerical optimization algorithms \cite{nocedal1999numerical}. Similarly, they often tend to contain logic that imposes restrictions for vectorized simulation and domain randomization of simulation geometry, which are crucial in robotics applications such as sampling-based model predictive control \cite{howell2022, kurtz2024hydrax} or sim-to-real transfer of reinforcement learning policies \cite{allshire2022transferring}. 

Collision detection routines of the state-of-the-art simulation approaches in computer graphics, such as barrier-based methods \cite{Li2020IPC}, pursue a different approach. These methods operate by explicitly considering contacts between vertex-face (V-F) and edge-edge (E-E) pairs in a smoothly differentiable manner, with the downside of being slower and less memory efficient. They also only generate unsigned penetration metrics and unsigned contact normals, whereas complementary-based contact dynamics formulations \cite{anitescu1997formulating, anitescu2004constraint, todorov2014convex, castro_sim} commonly used in robotics explicitly require signed versions.

The main contribution of this paper is to adapt the collision detection concepts of barrier-based methods for robotics by proposing a novel contact manifold generation method that:
i)~is smoothly differentiable to facilitate optimization, ii)~is fast and memory efficient to facilitate vectorization, iii)~generates signed penetration metrics and contact normals to be compatible with the complementarity-based methods for contact dynamics.



\section{Related Work} \label{sec:related_work}
The standard process employed by common simulators used in robotics \cite{todorov2012mujoco, drake, coumans2016pybullet} to generate a contact manifold (i.e., a properly distributed set of contact points) between two arbitrary surfaces can be abstracted into the following steps \cite{erleben2018methodology, gregorius2015robust}:
\begin{enumerate}[leftmargin=*, label={\color{ourblue}(\roman*)}]
    \item Converting each non-convex surface into a mesh with well-distributed facets via a computational geometry algorithm \cite{jakob2015instant, Corman:2025:RSP, oh2025pamo}, and decomposing it into a set of convex meshes \cite{wei2022coacd}. All of the remaining steps operate on pairs of these convex meshes.
    \item Running standard collision detection routines like GJK+EPA \cite{gilbert1988fast, montaut2022collision, van2001proximity} or SAT \cite{gregorius2013sat} to obtain information such as a pair of \say{witness points} (the end-points of the largest penetration distance) or closest mesh facets.
    \item Building a contact manifold that sufficiently represents the entire volume/area of intersection between surfaces.
\end{enumerate}
Alternative ways of implementing the last step are \cite{gregorius2015robust}:
\begin{itemize}[leftmargin=*, label={\color{ourblue}\textbf{\textbullet}}]
    \item Incrementally building it across simulation time-steps by maintaining a buffer of witness points based on distance heuristics about when to add or remove from the buffer.
    \item Building a contact manifold from scratch every time-step (i.e. \say{one-shot}) by: i)~running GJK+EPA to identify the penetration vector and tangential axes to it, and ii)~inducing small rotations on the two geometries multiple times around the tangential axes and re-running GJK+EPA each time.
    \item Building a contact manifold one-shot by applying polygon clipping algorithms \cite{sutherland1974reentrant} to the closest mesh-facets.
\end{itemize}
As expected, these routines involve discontinuities due to significant branching, which means simply applying automatic differentiation to them results in uninformative gradients, and it is necessary to devise further smoothing schemes. The work by \citet{montaut2023differentiable} proposes a modification to the standard GJK+EPA routine that uses randomized smoothing to obtain useful gradients via the score function estimator \cite{williams1992simple}. 
The main downsides of this approach are that: i)~it is sampling-based (and hence inefficient), ii)~it is not entirely transparent how exactly the parameters of the sampling distribution alter the overall dynamics, iii)~it only addresses witness point computation. The work of \citet{tracy2023differentiable} in turn takes a different approach and solves for the growth distance \cite{ong1996growth} between pairs of a representative set of convex primitives (including convex meshes) by formulating them as cone-constrained convex programs. Gradients through this convex program are then obtained via implicit differentiation and smoothed via modulating the log-barrier coefficient of the underlying interior-point solver \cite{tracy2024differentiability}. Its main downsides are: i)~it does not scale (in terms of speed and memory footprint) to meshes with a large number of faces, and ii)~it again only addresses witness point computation. 
In contrast, this paper proposes a deterministic method for generating a complete contact manifold that scales to complex meshes, while ensuring that each smoothing parameter influences the dynamics through a closed-form expression, thereby making its impact explicit.

Barrier-based methods for contact simulation \cite{Li2020IPC} take a different approach from the previously described standard routines. Instead of producing a contact manifold consisting of a relatively small number of points with signed distances and normals, they solely compute unsigned distances between V-F and E-E pairs. These are then used within a log-barrier term that enforces non-penetration constraints, added to the usual optimization objective for the implicit time integration of unconstrained rigid-body systems. While technically differentiable, there are two aspects of this approach incompatible with common simulation routines employed in robotics: i)~they do not produce signed distances and normals, ii)~they need acceleration structures (e.g. kd-trees or bounding volume hierarchies) to avoid the $O(\vertexnum * \facenum + \edgenum * \edgenum)$ cost for all V-F and E-E pair checks, which hinder the vectorization and JIT compilation utilities of common array computing frameworks \cite{jax}. This paper aims to address these bottlenecks. 

It is also beneficial to examine the collision detection routines of existing differentiable simulators. The Nimble simulator \cite{werling2021fast} implements a custom mesh-mesh collision routine that computes gradients analytically via a branching algorithm in the backward pass, which introduces discontinuities \cite{nimble_col}. The Dojo framework \cite{howell2022dojo} in turn only supports collisions between a sphere and a plane/sphere/capsule/box \cite{dojo_col}, which can be resolved via a single contact point and differentiated analytically. The MJX framework \cite{mjx} supports mesh-mesh collisions via the following routine \cite{gregorius2015robust}: i)~run SAT to identify edges and faces closest to contact, ii)~apply a clipping algorithm to build a polygonal contact manifold, iii)~use a heuristic to select the four points on the polygon with approximately maximal area. While the implementation replaces all branching control flow with equivalent branchless functionality provided by the $\texttt{jax.lax}$ module (to enable vectorization and JIT compilation), the discontinuities remain. The DiffMJX simulator \cite{paulus2025hard} in turn implements a version of the MJX collision routine with smoothed gradients via the \texttt{softjax} library \cite{Softjax2025}, but the smooth selection operations that enable this also cause a significant slowdown for gradient computation. These observations suggest that contact manifold construction presents a significant bottleneck in the differentiable simulation literature, a problem that this paper addresses. 

\section{Background: Smooth Approximations to Non-Smooth Operators} \label{sec:smooth_operators}
The proposed framework makes extensive use of existing literature around smooth approximations of operators such as comparisons, clipping, and top-K selection. The goal of this section is to provide a compressed summary of the related background and notation for ease of reference. While more detailed surveys can be referred to if needed \cite{Softjax2025}, the main point to emphasize is that whenever any non-smooth operation of a type covered in this section is employed later in the exposition, a smooth and differentiable counterpart exists (and is utilized by the software implementation). Throughout the section, $\softness$ denotes a positive smoothing factor (and approximations get more accurate but less smooth as $\softness \to 0$). The sigmoid function $\sigma(x) = \left(1 + \exp(-x)\right)^{-1}$ can be used to smoothly approximate a comparison operator $\llbracket x > a \rrbracket: \mathbb{R} \to [0, 1]$ (using Iverson-Bracket) via $\sigma((x-a)/\softness)$. The hyperbolic tangent function $\text{tanh}(x) = \sigma(2x) - 1$ can in turn be used to approximate the sign operator as $\text{tanh}(x/\softness)$. 
The $\text{relu}(x)$ function can be smoothly approximated using a softplus function $\text{s}_+(x) = \softness \log(1 + \exp(x / \softness))$. The clipping operation $f_\text{clip}(x, x_\text{min}, x_\text{max}) = x_\text{min} + \text{relu}(x - x_\text{min}) - \text{relu}(x - x_\text{max})$ can in turn be softened into a softclip operation $s_\text{clip}(x, x_\text{min}, x_\text{max})$ by replacing $\text{relu}$ operations with $\text{s}_+$. The scalar sigmoid and softplus functions can be generalized to operate on vectors $\bm x \in \mathbb{R}^D$ to obtain the softmax and realsoftmax (i.e. logsumexp) functions respectively, notated as $s_\text{argmax}(\bm x)_i = \exp(x_i / \softness)/\sum_{j=1}^D \exp(x_j / \softness)$ and $\text{LSE}(\bm x) = \softness\log[\sum_{i=1}^D\exp(x_i / \softness)]$. Note that $\bm s_\text{argmax}(\bm x)$ is a vector of probability masses that approximates an argmax operation, whereas $\text{LSE}(\bm x)$ is a scalar that corresponds to a soft selection of the maximum component of $\bm x$. 
For any vector $\bm x \in \mathbb{R}^D$, the top-K operator can be smoothed to obtain a matrix-valued soft top-K operation $\bm\Gamma_K(\bm{x}) \in [0, 1]^{K \times D}$, where the entry $\Gamma_K(\bm{x})_{ki} \approx 1$ if $x_i$ is the $k$-th largest entry of $\bm{x}$, and $\Gamma_K(\bm{x})_{ki} \approx 0$ otherwise. There exists a variety of such soft top-K operators \cite{prillo2020softsort, softK, grover2019stochastic, Softjax2025}, and this paper employs \cite{prillo2020softsort}. The matrix $\bm\Gamma_K(\bm{x})$ is a probabilistic representation of the \emph{indices} that correspond to the top-K entries, and these entries \emph{themselves} can be selected simply via a matrix product $\bm\Gamma_K(\bm{x}) \bm x \in \mathbb{R}^K$.

\begin{figure}
\begin{center}
\includegraphics[width=0.9\columnwidth]{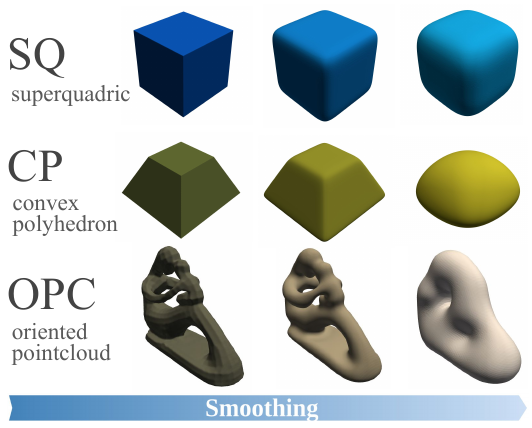}
\end{center}
\vspace{-2mm}
\caption{Visualizations of how the proposed SDF primitives behave with increasing levels of smoothing.}\label{fig:SDF_smoothing}
\vspace{-5mm}
\end{figure}

\section{Methods}
\subsection{A Birds-Eye View of the Most Essential Steps} \label{sec:high-level}
The presentation of the method follows a top-down exposition, with this section providing an overview of how all contributions fit together and the subsequent sections elaborating on each. 
Let $(\surface_1, \surface_2)$ be two (not necessarily convex) closed surfaces in 3D, with $SE(3)$ poses $\pose_1, \pose_2 \in \mathbb{R}^{4 \times 4}$ that can be smoothly parametrized via 6D $SE(3)$ vectors $\bm s_{1} = \log(\pose_1), \, \bm s_{2} = \log(\pose_2)$. 
The goal of contact manifold generation is to implement a function $g_{1,2} : \mathbb{R}^{6+6} \to \mathbb{R}^{\contactpointnum\times (3+1+3)} \ ; \ g_{1,2}(\bm s_1, \bm s_2) = \{(\mathbf p_\contactpointidx, d_\contactpointidx, \mathbf n_\contactpointidx)\}_{\contactpointidx=1}^{\contactpointnum}$ that outputs $\contactpointnum$ contact points $\bm p_k \in \mathbb{R}^3$, signed distances $d_\contactpointidx \in \mathbb{R}$, and contact normals $\normal_\contactpointidx \in \mathbb{R}^3$. The task at hand is to define $g_{1,2}$ such that it is smoothly differentiable and efficiently vectorizable.
The proposed framework achieves this as follows:\footnote{Since forward kinematics is already smoothly differentiable and is not the focus of this paper, the rest of the exposition omits the dependencies on $\bm s_1, \bm s_2$ for brevity and all quantities will be assumed to be expressed in world coordinates after forward kinematics.}
\begin{itemize}[leftmargin=*, label={\color{ourblue}\textbf{\textbullet}}]
    \item Every surface has two representations associated with it: i)~a mesh (with vertices $\vertexset = \{\mathbf v_\vertexidx \in \mathbb R^3\}$ and edges $\edgeset = \{\mathbf{e}_\edgeidx \in \mathbb{Z}^2\}$), and ii)~a smoothly differentiable SDF $\phi: \mathbb{R}^3\to \mathbb{R}$.
    \item All vertices (on both surfaces) are assigned penetration metrics using the SDF of the opposing surface. The proposed framework provides methods to build such SDFs efficiently from a set of analytical primitives $\primitiveset$ (where $\primitivenum$ is much smaller than $\vertexnum$ and $\facenum$), making this a relatively cheap ${O}(\vertexnum \times \primitivenum)$ operation. All edges are then assigned penetration metrics based on a weighted sum of the penetration metrics of their corner vertices, which is again relatively cheap (${O}(\edgenum \times 2)$). Then $(\vertexnum[1],\vertexnum[2])$ vertices and $(\edgenum[1], \edgenum[2])$ edges on each surface that are in the deepest penetration are selected via soft top-K operations. The numbers $\vertexnum[\binaryidx]$ and $\edgenum[\binaryidx]$ for $\binaryidx\in\{1, 2\}$ are set as part of the simulation description and kept fixed throughout the lifetime of the simulation. This design choice, same as MJX \cite{mjx}, is because JIT compilation and vectorization require the number of contact points to remain constant.
    This step essentially implements a differentiable acceleration structure.
    \item As discussed in \citet{Li2020IPC}, considering V-F and E-E contact types is sufficient to transcribe constraints that prevent penetrations between two surfaces in arbitrary relative poses. In the proposed framework, V-F contacts are replaced with vertex-SDF (V-S) contacts for efficiency. 
    For each of the $\edgenum[1] \times \edgenum[2]$ edge-edge pairs, two contact points (one on each edge) are generated together with the associated $\signeddistance_\contactpointidx$ and $\normal_\contactpointidx$ (as well as a scalar \say{activity indicator} $\gamma$ that will be introduced later) via a novel E-E contact routine.
\end{itemize}

\subsection{Building Smooth SDF Approximations of Arbitrary Surfaces for Implementing Vertex-SDF Contacts}\label{sec:sdf_prim}

As shown in \cref{fig:main} and \ref{fig:SDF_smoothing}, smoothly differentiable SDFs are built from combining analytical primitives $\phi_\primitiveidx(\bm x): \mathbb{R}^3 \to \mathbb{R}$ via smooth union and subtraction operations \cite{sdf_algebra, smoothmin} implemented using the logsumexp (real-softmax) function. In particular:
\begin{itemize}[leftmargin=*, label={\color{ourblue}\textbf{\textbullet}}]
    \item The smooth union of $\primitivenum$ primitives $\{\phi_\primitiveidx(\bm x)\}_{\primitiveidx=1}^{\primitivenum}$ is given by $\phi(\bm x) = -\text{LSE}([-\phi_1(\bm x),\cdots, -\phi_{\primitivenum}(\bm x)])$.
    \item The smooth subtraction of $\phi_{-}(\bm x)$ from $\phi_{+}(\bm x)$ is given by $\phi(\bm x) = \text{LSE}(\phi_{+}(\bm x), -\phi_{-}(\bm x))$.
\end{itemize}
This approach of fusing analytical primitives to obtain implicit surface descriptions has long been used by the computer graphics community to model complex surfaces in great detail~\cite{shadertoy, quileztalk}. As one cannot simply code new implicit surface descriptions from scratch for every new robotics simulation, the proposed framework provides three families of primitives. These are (in the order of most to least efficient in terms of memory and compute requirements): i)~superquadrics, ii)~convex polyhedra, and iii)~oriented pointclouds.

\subsubsection{Superquadric (SQ) Primitives}
An SQ in its canonical form (i.e. centered at the origin, axis-aligned, with unit scaling) is defined by an \emph{inside--outside} implicit function $f:\mathbb{R}^3\rightarrow\mathbb{R}$ such that the surface corresponds to the unit level set $f(\bm x)=1$ (exterior: $f(\bm x)>1$).
The standard definition of $f$ and the corresponding approximate signed distance $\phi$ are \cite{liu2023marching}:
\begin{align}
    \bm x &= [x, y, z] \\
    f(\bm x) &= \left[(x^2)^{1/\epsilon_2} + (y^2)^{1/\epsilon_2}\right]^{\epsilon_2 / \epsilon_1} + (z^2)^{1/\epsilon_1} \\
    \phi(\bm x) &= (1 - f(\bm x)^{-\epsilon_1/2}) / \|\bm x\|\,,
\end{align}
where $\epsilon_{\{1,2\}}$ control the ``boxiness''.
Shape scaling can be obtained by normalizing the coordinates $\tilde{\bm x}=[x/a,\;y/b,\;z/c]$ with axis lengths $(a,b,c)$.
Even though $\phi(\bm x)=0$ holds on the SQ surface and $\nabla \phi(\bm x)$ points in the correct direction everywhere, the approximation does not satisfy the Eikonal property $|\nabla \phi(\bm x)| = 1$ far away from the SQ surface, meaning the value $\phi(\bm x)$ represents a metrically accurate distance only near the SQ surface. This does not pose a problem for simulation for two reasons: i)~contact points always remain very close to the SQ surface (since contact forces prevent further penetration), ii)~the complementarity-based formulations of contact dynamics can accommodate nonmetrical $\signeddistance_\vertexidx$ as long as its zero crossing happens on the surface (since the non-metric scaling simply acts as a virtual contact impedance). Signed distances and normals are therefore obtained from $\signeddistance_\vertexidx = \phi(\bm x)$ and $\normal_\vertexidx = \nabla f(\bm x) / \sqrt{\softness + \|\nabla f(\bm x)\|^2}$.

\subsubsection{Convex Polyhedron (CP) Primitives}

A convex polyhedron is a collection of planes $\{(\normal_\polyhedronidx, \point_\polyhedronidx)\}_{\polyhedronidx=1}^\polyhedronnum$, where $\normal_\polyhedronidx$ and $\point_\polyhedronidx$ denote the normal of the plane and a point on it. The exact SDF and its continuously differentiable approximation are \cite{yang2024contactsdf}:
\begin{align}
    \phi_{\text{exact}}(\bm x) & = \max_\polyhedronidx \ \normal_\polyhedronidx^\trsp (\bm x - \point_\polyhedronidx) \\
    \phi(\bm x) &= \text{LSE}([\normal_1^\trsp (\bm x - \point_1), \cdots, \normal_\polyhedronnum^\trsp (\bm x - \point_\polyhedronnum])
\end{align}
Contact normals are then obtained as $\nabla \phi(\bm x) / \sqrt{\softness + \nabla \|\phi(\bm x)\|^2}$.

\subsubsection{Oriented Pointcloud (OPC) Primitives}

If one has access to an oriented pointcloud $\{(\point_\opcidx, \normal_\opcidx)\}_{\opcidx=1}^\opcnum$ on any non-convex surface, a radial basis interpolation of the following form can be used to obtain a smooth SDF approximation \cite{beker2025}:
\begin{equation}
    \phi(\bm x) = \frac{\sum_{\opcidx=1}^\opcnum \bm B_\opcidx(\bm x - \point_\opcidx \ ; \ \bm \theta_\opcidx) \ \normal_\opcidx^\trsp (\bm x - \point_\opcidx)}{\sum_\opcidx \bm B_\opcidx(\bm x - \point_\opcidx \ ; \ \bm \theta_\opcidx)}
\end{equation}
Contact normals are again obtained as $\nabla \phi(\bm x) / \sqrt{\softness + \|\nabla \phi(\bm x)\|^2}$.

In general, SQ primitives are very compute- and memory-efficient and can model a wide-range of complex geometries~\cite{barr1981superquadrics}. Therefore, the default option in the proposed framework is to use the algorithm of \citet{liu2023marching} to decompose collision geometries into SQ primitives. CP primitives are less efficient, but can model sharp corners better and can generally be added to the decomposition whenever an SQ primitive would be insufficient to model a feature. OPC primitives are the most expressive and can model arbitrary nonconvex surfaces, but are also the least efficient. They should only be used when a feature to be modeled is inherently non-convex, and SQ or CP primitives would be insufficient. 


\subsection{Generating Vertex-SDF (V-S) Contacts}\label{sec:v-s}
For every selected vertex $\vertex_\vertexidx$, a V-S contact is generated by utilizing the smooth SDF approximation $\phi$ of the opposing surface to compute the associated signed distance $\signeddistance_\vertexidx = \phi(\vertex_\vertexidx)$ and contact normal $\normal_\vertexidx = \nabla \phi(\vertex_\vertexidx)$.\footnote{Note that for SQ primitives, normals are actually obtained from $\nabla f$ and not $\nabla \phi$, since the latter is non-monotonic and can cause inverted normals.} Since $\phi$ is smoothly differentiable, so are $\signeddistance_\vertexidx$ and  $\normal_\vertexidx$.

\subsection{Generating Edge-Edge (E-E) Contacts}\label{sec:e-e}
\subsubsection{Witness Point Computation (General Case)} \label{sec:general_wp}
Points $\point$ on a convex polygon with
$N$ corners $\vertex_{1}, \cdots, \vertex_N$ can be parameterized with a linear term 
$\point(\bm{\alpha}) = \sum_\vertexidx \vertex_{\vertexidx}\alpha_{\vertexidx}$ and an 
equality constraint $\sum_\vertexidx \alpha_{\vertexidx} = 1$. 
Therefore witness points $\point_1(\astar_1),\point_2(\astar_2)$ between two arbitrary polygons $(1,2)$ can be found via a QP:
\begin{align}
\min_{\boldsymbol{\alpha}_1 \in \mathbb{R}^{N_1},\; \boldsymbol{\alpha}_2 \in \mathbb{R}^{N_2}}\label{eq:wp_qp}
&\; \left(\boldsymbol{p}_1(\boldsymbol{\alpha}_1) - \boldsymbol{p}_2(\boldsymbol{\alpha}_2)\right)^\trsp
\left(\boldsymbol{p}_1(\boldsymbol{\alpha}_1) - \boldsymbol{p}_2(\boldsymbol{\alpha}_2)\right) \\
\text{s.t.}\quad
&\sum_{1\leq i<N_1} \alpha_{1,i} = 1 \: , \: \bm \alpha_1 \succeq \bm 0 \\
& \sum_{1\leq i<N_2} \alpha_{2,i} = 1 \: , \: \bm \alpha_2 \succeq \bm 0\,.
\end{align}
Since mesh primitives such as vertices ($N=1$), edges ($N=2$), and triangles ($N=3$) are all convex polygons, witness point computations between them can all be solved as QPs.

Jumps in the solution of such a QP can happen due to two reasons: i)~the quadratic term loses rank (i.e., becomes only positive semi-definite) and there are infinitely many solutions
, or ii)~inequality constraints are switching on and off. The most common way to smooth (i) is to add L$_2$ (i.e., Tikhonov) regularization \cite{nocedal1999numerical}, and (ii)~is to use an interior point method and increase the weight of the log barrier function \cite{tracy2024differentiability}. The problem with this log barrier approach is that it involves an iterative numerical scheme (e.g.,  primal-dual Newton iterations on the KKT system) to find the \emph{exact} minimum on the log barrier, which is an overkill for small QPs (e.g., E-E and V-F collisions), considering that the non-smoothed versions can be solved via active set methods \cite{nocedal1999numerical} much faster and with a fraction of the effort. 
Moreover, such iterative procedures are inefficient for vectorization, because a vectorized iterative solver applied to a batch of QPs will only finish as quickly as the slowest-converging QP in the batch.
Therefore, for the particular class of QPs we want to solve (i.e. witness point computations between V-E-F primitives), it is more promising to search for ways to alter active set methods to make them smoothly differentiable and efficiently vectorizable.

The main bottleneck with employing the active set method is that it involves: i)~branching control flow (i.e., when identifying which subset of the constraints are active), and ii)~non-smooth logic operators. The proposed framework overcomes these by implementing smoothed variants of the following most trivial, analytical (i.e., non-iterative), branchless active set method: 
\begin{enumerate}[leftmargin=*, label={\color{ourblue}(\roman*)}]
    \item Analytically compute the unconstrained minimizer $\bm x^{u\star}$.
    \item Analytically compute the constrained minimizers $\{\bm x_i^{c\star}\}_{i=1}^{2^N}$ for all $2^N$ possible ways of turning the $N$ inequality constraints $g(\bm x)_i \geq 0$ into equality constraints by switching them on/off, and pick the member from $\{\bm x_{i}^{c\star}\}_{i=1}^{2^N}$ that minimizes the objective as $\bm x^{c\star}$.
    \item Compute a binary indicator $\gamma = \prod_i [g(\bm x^{u\star})_i \geq 0]$ about whether $\bm x^{u\star}$ is within the constraint set.
    \item Return $\bm x^\star = \gamma \bm x^{u\star} + (1 - \gamma) \bm x^{c\star}$.
\end{enumerate}
Clearly, this is only scalable for very small QPs, and as the QP size increases, the interior-point method will quickly become more efficient. Yet, for simple witness point routines (e.g., V-F, E-E), we have found this to be significantly faster, especially under massive vectorization. It can also be easily smoothed using approximations from S.\ref{sec:smooth_operators} for operations such as inequality checks and $\argmin$ evaluations. The proposed framework implements variants of this approach for both V-F and E-E contacts, although as previously discussed in S.\ref{sec:high-level}, the V-F routine is never actually used since V-S contacts are more efficient. The supplementary document contains exact descriptions for both V-F and E-E routines in pseudo-code.

\begin{figure}[t]
    \centering
    \includegraphics[width=\linewidth]{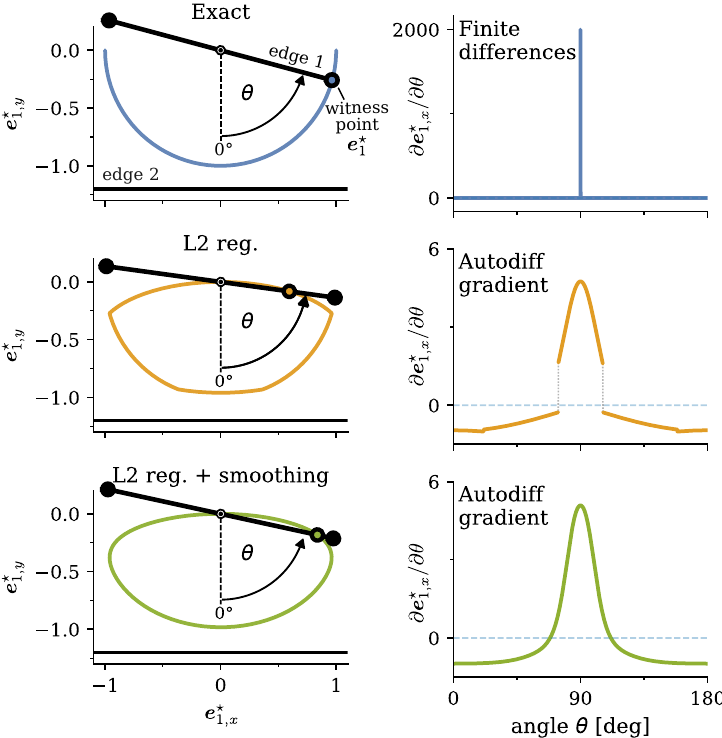}
    \caption{(\textbf{Left}) A didactic experiment illustrating how E-E witness points behave under the proposed smoothing scheme. There are two edges lying on the $xy$ plane, edge-2 being fixed and extending in the $x$ direction with $y=-1.2$, and edge-1 rotating around the $z$ axis of the origin. The colored lines plot the positions traced by the witness point $\bm{e}_{1}^{\star}(\theta) = [\bm{e}_{1,x}^{\star}, \bm{e}_{1,y}^{\star}, 0]$ on edge-1 as it completes a half rotation $\theta: 0 \to \pi$. 
    Without smoothing (top left), $\bm{e}_{1}^{\star}(\theta)$ exhibits a jump as the system passes through the pathological parallel configuration $\theta = \pi/2$. The center and bottom plots visualize the smoothing contributions of L$_2$ regularization and employing the approximations from \cref{sec:smooth_operators}. (\textbf{Right}) The gradients of $\bm{e}_{1}^{\star}(\theta)$ for the three cases above. It can be seen that in the absence of any smoothing, the finite difference gradients are uninformative (e.g., very large near the pathological configuration), which is fixed by the proposed smoothing scheme.}
    \label{fig:edge_edge_contact_point}
    \vspace{-5mm}
\end{figure}
\subsubsection{Creating E-E Witness Points} \label{sec:e_e_wp}
Points $\bm e$ on an edge between corners $\vertex_I, \vertex_{II} \in \mathbb{R}^3$ can be parameterized with a linear term $\edge(\alpha) = \vertex_I + (\vertex_{II} - \vertex_I) \alpha$ and a box constraint $\alpha \in [0, 1]$. This means for E-E contacts, the unsigned distance QP \eqref{eq:wp_qp} can be expressed using only box-constraints:
\begin{align}
  \astar  = \ \argmin_{\alpha_1,\alpha_2 \in \mathbb{R}} \: & \left(\edge_1(\alpha_1) - \edge_2(\alpha_2) \right)^\trsp \left(\edge_1(\alpha_1) - \edge_2(\alpha_2) \right) \nonumber\\
    \ \text{s.t. } \: & 0 \leq \alpha_1,\alpha_2 \leq 1,
    \label{eq:e-e_QP}
\end{align}
with $\astar = [\alpha_1^{\star}, \alpha_2^{\star}]$.

\paragraph*{L$_2$ regularization} 
As shown in Fig.\ref{fig:edge_edge_contact_point}, when the tangential directions $\bm t = \vertex_{II} - \vertex_I$ of two edges point in the exact same 3D direction, the quadratic term in the QP above is rank deficient, and the minimum is not unique, which means the witness points exhibit a jump around this pathological configuration. 
As discussed previously, an L$_2$ regularization term $R(\alpha_1,\alpha_2)=\lambda \left[(\alpha_1 - 0.5)^2 + (\alpha_2 - 0.5)^2\right]$ is added to the objective of \eqref{eq:e-e_QP} to counteract this. This biases the solution towards the center of the edge, with increasing $\lambda$ resulting in more smoothing (\cref{fig:edge_edge_contact_point} - middle). Then the QP is solved with the following variant of the general analytical active set approach from S.\ref{sec:general_wp}:
\begin{enumerate}[leftmargin=*, label={\color{ourblue}(\roman*)}]
    \item Compute the unconstrained minimizer $\astaru = \bm Q^{-1}\bm c$.
    \item Compute the minimizers $\{\bm \astarc_1, \ldots, \astarc_4\}$ along all four edges of the box constraints. For example along the edge $[\alpha_1=1, \alpha_2]$ the minimizer is $\Big[1, f_\text{clip}\big(-\frac{Q_{12} - c_2}{Q_{22}}, 0, 1 \big)\big]$, smoothed via a softclip operation. Then select the minimizer of the objective from $\{\bm \astarc_1, \ldots, \astarc_4\}$ as $\astarc$, smoothed via a softmax operation.
    \item Compute the indicator ${^\text{con}\gamma} = [0 \leq \astarucomp_1 \leq 1] * [0 \leq \astarucomp_2 \leq 1]$, smoothed using soft comparison operators.
    \item Return $\astar = {^\text{con}\gamma} \, \astaru + (1 - {^\text{con}\gamma})\astarc$.
\end{enumerate}

Note that an E-E contact is \say{colliding} (i.e., moving the edges along $\bm t_1 \times \bm t_2$ can create a contact) only when $\astaru$ lies within the constraint set. Therefore, ${^\text{con}\gamma}$ can be used as a soft indicator to identify if a given E-E contact is colliding. 


\subsubsection{Assigning Signed Distances and Normals} \label{sec:sign_assignment}
After obtaining $\astar$, the corresponding witness points are obtained as $\edge_1^{\star}=\edge_1(\alpha_1^{\star})$ and $\edge_2^{\star}=\edge_2(\alpha_2^{\star})$, yielding $\Delta \edge = \edge_1^{\star} - \edge_2^{\star}$. 
Then, the unsigned distance and unsigned normal for the E-E pair are simply $\bar d = \|\Delta \edge\|$ and $\bar {\normal} = \Delta \edge/\sqrt{\|\Delta \edge\|^2 + \epsilon}$.\footnote{The normal $\bar {\bm n}$ can be constructed from either $\bm t_1 \times \bm t_2$ or $\Delta \edge$. The former cannot handle penetrations that happen when two edges are completely parallel (as $\bm t_1 \times \bm t_2 = \bm 0$), and the latter cannot handle the case when two edges are exactly intersecting, but there is no penetration yet as ${\Delta \edge} = \bm 0$. 
Because most commonly used simulators rely on penalty formulations that only create forces after a penetration is present, the latter is the more suitable choice. 
Note that as long as an E-E contact is colliding, non-parallel, and non-intersecting, $\bm t_1 \times \bm t_2$ and $\Delta \edge$ lie along the same line and are hence equivalent.} 

\begin{figure}[t]
    \centering
    \includegraphics[width=0.95\linewidth]{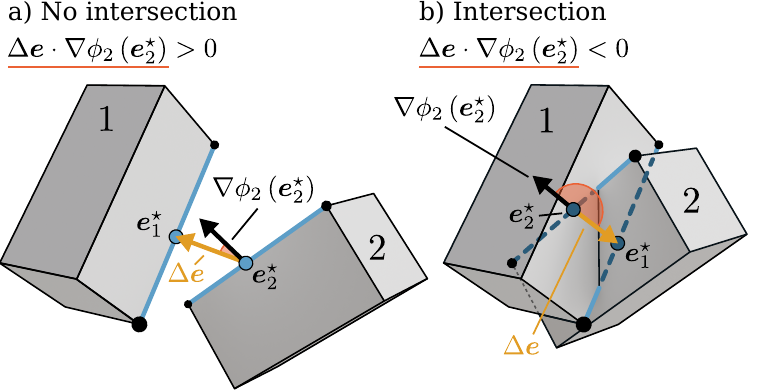}
    \caption{Creating E-E Contacts: The witness points $\bm{e}_i^{\star}$, as well as the corresponding unsigned distances and normals, are obtained from the (smoothed) analytical active set method described in S.\ref{sec:e_e_wp}. The subsequent sign assignment for distances and normals in turn utilizes the SDFs by evaluating the dot product of the unsigned normal direction and the SDF gradient for the penetrated surface evaluated at each witness point.}
    \label{fig:edge-edge:sign}
    \vspace{-2.5mm}
\end{figure}

\begin{figure}[th]
\begin{center}
\includegraphics[width=0.95\columnwidth]{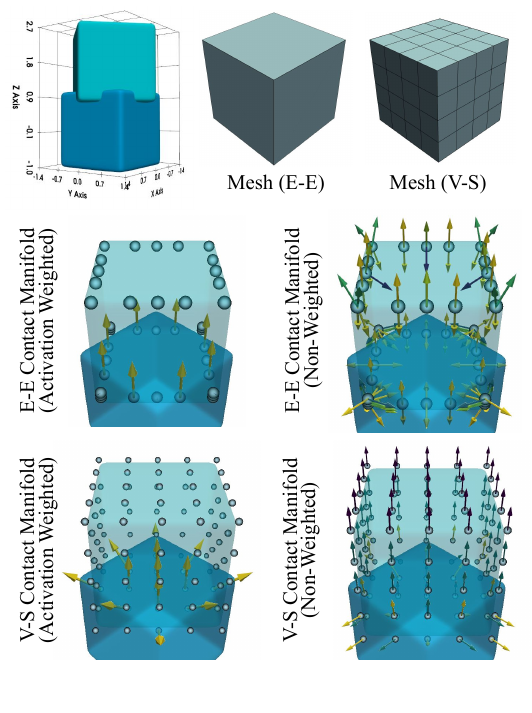}
\end{center}
\vspace{-9mm}
\caption{Contact manifold visualizations for box-box collisions. SDFs are obtained from SQ primitives ($\epsilon_1 = \epsilon_2 = 0.1$). For illustration clarity, E-E contacts use the simplest cube mesh with $8$ vertices, $12$ edges, and $6$ faces, whereas V-S contacts use a more densely meshed version of the same cube with $98$ vertices. In both cases, only half of the contact points (belonging to the top box) are visualized, resulting in two manifolds: one with $12 \times 12 = 144$ E-E contacts and one with $98$ V-S contacts. For both, two versions are shown: a version where normals are scaled with the activity indicator (to visualize which contacts are actually selected by the routine), and a version where they are unscaled. It can be seen that normals point in the right directions, and the correct points are selected.}
\label{fig:box-box}
\vspace{-5mm}
\end{figure}

\paragraph*{Determining the sign}
The next step is to assign signs to $\bar d$ and $\bar {\bm n}$ such that contact forces always oppose penetration. 
As shown in Fig. \ref{fig:box-box}, the main difficulty of assigning the sign is in handling cases where there are penetrations, but the E-E witness points lie exactly on the intersection of the two surfaces. 
The standard way to address this problem is to rely on the separating plane between the surfaces to determine the sign \cite{gregorius2015robust}. 
When one has access to SDFs, the analogous approach is to realize that the contact normal for witness point $\edge_1^{\star}$ should always point in the same direction as the normal $\normal_2 := \nabla \phi_2(\edge_2^{\star})$.
Hence, if we construct a separating plane using $\edge_2^{\star}$ and $\normal_2$, the correct sign is obtained as $\gamma_1 = \text{sign}(\normal_2^\trsp \bar{\bm n})$ (and vice-versa for $\gamma_2 = \text{sign}(\normal_1^\trsp \bar{\bm n})$), smoothed via a soft-sign. 
Therefore the signed distances and contact normals for both witness points $\edge_1^{\star}$ and $\edge_2^{\star}$ can be obtained as:
\begin{equation}
   d_{\binaryidx} = \gamma_{\binaryidx} \bar d \text{ and } \normal_{\binaryidx} = \gamma_{\binaryidx} \bar \normal \text{ for } \binaryidx\in\{0,1\}. 
\end{equation}

\paragraph*{Determining whether there is a penetration}
To determine if a witness point is actually penetrating, one can compute
\begin{align}
       ^\text{pen}\gamma_{1} &= \sigma(-\phi_2(\edge_{1}^{\star}) / \softness), \label{eq:sdf_check}
\end{align}
to approximate $[\phi_2(\edge_1^{\star}) \leq 0]$ via a soft comparison operator.

\paragraph*{Determining neighboring edges}
E-E contacts should only be active if they are neighbors.
Therefore, we need a mechanism that determines if edges are close to each other. After witness point computation, there are $|E_1| \times |E_2|$ unsigned distances $\bar d_{k, l}$ computed between all (ordered) E-E pairs, with $k \in \{1, \cdots, |E_1|\}$ and $l \in \{1, \cdots, |E_2|\}$. Let us arrange all $\bar d_{k, l}$ into a matrix $\bm D \in \mathbb{R}^{|E_1| \times |E_2|}$.
%
To check if an (ordered) E-E pair is a nearest neighbor, we can compute two soft indicators (since computing nearest neighbors is an asymmetric operation)
\begin{align}
    ^\text{nn}\Gamma_{1, k, l} &= s_{\text{argmax}}\Bigl(-\left[ \ D_{k,1} \: \cdots \: D_{k \, , \, |E_2|} \ \right]\Bigr)_l \label{eq:nn_check} \\
    ^\text{nn}\Gamma_{2, k, l} &= s_{\text{argmax}}\Bigl(-\left[ \ D_{1,l} \: \cdots \: D_{|E_1| \, , \, l} \ \right]\Bigr)_k 
\end{align}
by simply applying the soft argmin operation row- and column-wise on $\bm D$, to obtain ${^\text{nn}\bm{\Gamma}_{1}}, {^\text{nn}\bm{\Gamma}_{2}}  \in \mathbb{R}^{|E_1| \times |E_2|}$.

\paragraph*{Determining clashing edges}
E-E contacts should only be active if they are \say{clashing} (i.e. if they have opposing normals). This can be checked using the normals $\normal_1 = \nabla \phi_1(\edge_1^{\star})$ and $\normal_2 = \nabla \phi_2(\edge_2^{\star})$ evaluated at the witness points $\edge_1^{\star}, \edge_2^{\star}$ as:
\begin{align}
       ^\text{cl}\gamma_{1} &= \sigma \left( \, -\normal_1^\trsp \normal_2 / \tau \, \right) ,
\end{align}
approximating $[ \normal_1^\trsp \normal_2 \leq 0]$ again via a soft comparison operator.
 
\paragraph*{Determining which E-E collisions are active}
A witness point in an E-E pair should be an active contact point if:
\begin{itemize}[leftmargin=*, label={\color{ourblue}\textbf{\textbullet}}]
    \item The E-E pair is colliding.
    \item The witness point is penetrating into the opposing surface.
    \item The E-E pair constitutes nearest-neighbors.
    \item The E-E pair is clashing.
\end{itemize}

Much like the soft nearest-neighbor indicators ${^\text{nn}\Gamma_{1, k, l}}$ and ${^\text{nn}\Gamma_{2, k, l}}$; the soft indicators ${^\text{con}\gamma}$ for collision, $^\text{pen}\gamma$ for penetration, and $^\text{cl}\gamma$ for clashing are also defined for each of the two witness points in all $|E_1| \times |E_2|$ (ordered) E-E pairs. This means they can also be reshaped into a matrix format to obtain ${^\text{con}\bm{\Gamma}_1}, {^\text{con}\bm{\Gamma}_2}, {^\text{pen}\bm{\Gamma}_1}, {^\text{pen}\bm{\Gamma}_2}, {^\text{cl}\bm{\Gamma}_1}, {^\text{cl}\bm{\Gamma}_2}, \in \mathbb{R}^{|E_1| \times |E_2|}$. In the proposed framework, all of these soft indicators are then combined to yield an \say{activity indicator}
\begin{equation}
    \Gamma_{b, k, l} = {^\text{con}\Gamma_{b, k, l}} \, * \, {^\text{pen}\Gamma_{b, k, l}} \, * \, {^\text{nn}\Gamma_{b, k, l}} * {^\text{cl}\Gamma_{b, k, l}} \: \text{ for } \binaryidx\in\{0,1\}
\end{equation}
for all E-E contacts.
The subsequent contact dynamics routine can then use this information to turn contacts on or off.\footnote{For example in a spring-damper model one can scale the stiffness with $\Gamma$, or analogously in the primal cone-complementarity formulation \cite{todorov2014convex, castro_sim} used by Mujoco \cite{todorov2012mujoco} and Drake \cite{drake}, the contact impedance can again be scaled.} 

\section{Experiments}

The aim of this section is to validate that the proposed routine can provide a contact manifold that adequately represents the volume of intersection between colliding surfaces in a manner that is differentiable, fast, and massively vectorizable.

\subsubsection{Validity of the contact manifold}\label{sec:validity_exp}
Two boxes resting on each other is one of the simplest cases in collision detection where witness point routines like GJK+EPA alone are not sufficient, and it is necessary to build a proper contact manifold (of up to 8 contact points \cite{todorov2012mujoco}) with both V-F and E-E types for robust simulation. It is also a sufficiently representative case since non-convex meshes can be thought of as unions of convex polyhedra, thus making it the quintessential example covered by almost all pedagogical treatments of contact manifold generation \cite{gregorius2015robust}. \Cref{fig:box-box} also uses this example to visualize what contact manifolds generated by the proposed routine look like. As can be seen, the contact normals point in the correct directions and the contact activity indicators properly filter any contributions from non-active E-E and V-S pairs.

\begin{figure}[th]
\begin{center}
\includegraphics[width=\columnwidth]{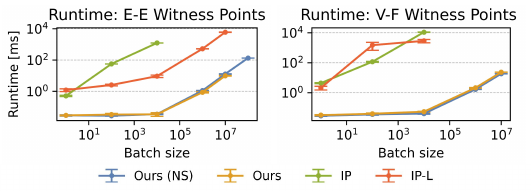}
\end{center}
\vspace{-2mm}
\caption{Plots showing the speed and vectorizability of the proposed active-set approach for witness point computation (Ours, and the non-smoothed version Ours-NS), in comparison with log-barrier smoothing approaches. The latter are implemented via two interior point solvers, primal-dual Newton iterations on the KKT system (IP), and L-BFGS iterations with a log barrier penalty (IP-L). It can be seen that the proposed routines are fast and efficiently vectorizable.}
\label{fig:wp_exp}
\vspace{-5mm}
\end{figure}
\subsubsection{Speed and Vectorizability of V-F, V-S, and E-E Collisions}\label{sec:V_S_F_E_exp}
\Cref{fig:wp_exp} shows how the runtimes of the proposed V-F and E-E witness point routines change as one increases the number of vectorized queries from $1$ to $10^7$. Queries are obtained by sampling V-F and E-E pairs from the uniform distribution over the unit cube, and the reported numbers are the mean and standard deviations from 100 trials. The baselines are:
\begin{itemize}[leftmargin=*, label={\color{ourblue}\textbf{\textbullet}}]
    \item \ul{Ours-NS} (no smoothing): Solving the V-F and E-E QPs via our proposed analytical active set routine, only without replacing non-smooth operations (e.g., comparisons and argmin operations) with their smooth approximations.
    \item \ul{IP}: Solving the V-F and E-E QPs via the primal-dual interior point method (as implemented by the \texttt{qpax} library \cite{tracy2024differentiability}), and smoothing them by changing the log-barrier coefficient.
    \item \ul{IP-L}: Solving the exact same optimization objective (i.e. with the same log-barrier penalty) as the IP baseline, but using L-BFGS \cite{byrd1995limited} for efficiency (as implemented by the \texttt{jaxopt} library \cite{jaxopt_implicit_diff}) rather than the the primal-dual interior point method. For V-F contacts, this requires parameterizing points on mesh faces via bilinear interpolation to ensure that there are only box constraints (i.e. to be able to apply LBFGS).
\end{itemize}
It can be seen that the proposed analytical smoothing approach is orders of magnitude faster and more suitable for vectorization compared to the standard log-barrier approach for smoothing inequality constrained optimization problems. 

\begin{figure}[th]
\begin{center}
\includegraphics[width=\columnwidth]{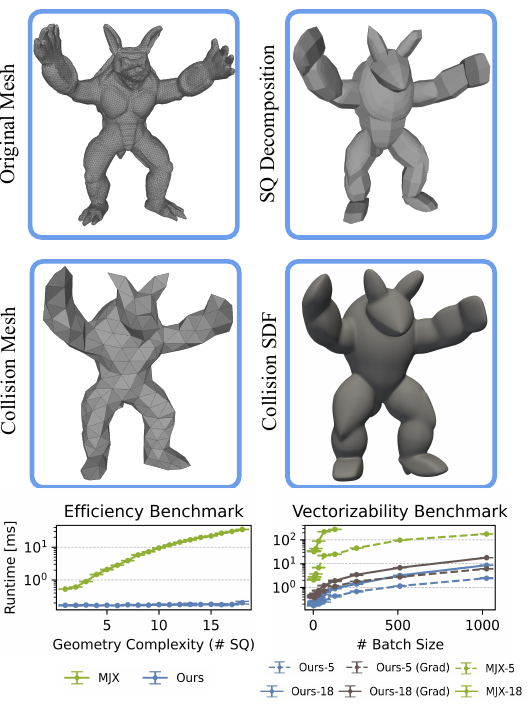}
\end{center}
\vspace{-3mm}
\caption{Plots showing the speed and vectorizability of the proposed contact manifold generation routine, compared to the analogous routine of the well-established MJX simulator. The setup involves collisions between two non-convex armadillo geometries across 100 random configurations per-trial. Surfaces are represented with up to 18 SQs each, which is a controlled variable characterizing the effects of geometry complexity on the runtime. The second controlled variable is the vectorization batch size. 
Notice the logarithmic scale and the orders of magnitude improvement
compared to MJX.}
\label{fig:bench_exp}
\vspace{-5mm}
\end{figure}
\subsubsection{Speed and Vectorizability of the Entire Routine}\label{sec:routine_exp} This section compares the speed of the proposed contact manifold generation routine and that of the MJX simulator \cite{mjx}, in a setting that involves two nonconvex meshes of an armadillo. For the proposed routine, the smooth SDF representation for the original armadillo mesh is constructed from 18 SQ primitives whose parameters are obtained using the decomposition method of \citet{liu2023marching} (\cref{fig:bench_exp}). The associated mesh representation (used by the proposed method) consists of 305 vertices, 605 faces, and 908 unique edges and is obtained by applying marching cubes to the smooth SDF representation and then applying the retopology method of \cite{jakob2015instant}. 
There is no soft top-K selection on vertices, creating $305 \times 2 = 610$ potential V-S contacts. For E-E contacts, soft top-K selection with $K=18$ is used (i.e., one edge per SQ), resulting in $18\times18\times2 = 648$ potential E-E contacts. To obtain collision meshes for the MJX routine, the 18 SQ primitives are converted to coarse convex meshes (i.e., since MJX requires all mesh geometries to be convex with fewer than 32 vertices) \footnote{We have also tried applying CoACD \cite{wei2022coacd} to the same mesh representation (with 305 vertices) used for the proposed routine to directly obtain a convex decomposition that can then be passed to MJX. This was significantly slower than using 18 coarse SQ meshes.}, resulting in $18\times18$ convex-convex collision pairs and hence $18\times18 \times 4 = 1296$ potential contact points.

The first plot of \cref{fig:bench_exp} shows how the runtimes of the two routines change depending on geometry complexity in a non-vectorized setting as the armadillo geometry is gradually reconstructed from $(1\times 1) \to (18 \times 18)$ SQ primitives. The reported numbers are again means and standard deviations over 100 trials, where the relative 6D poses of the two armadillos are randomized. It can be seen that the proposed framework is faster. The second plot of \cref{fig:bench_exp} runs the same experiment in a setting where the number of vectorized queries changes from $1 \to 1024$, only using the two cases with $5\times 5$ and $18 \times 18$ SQ primitives in terms of geometry complexity for brevity. The experiment suggests that the proposed framework is significantly more suitable for massive vectorization. For this same experimental setup, the speed and vectorizability of gradient computations in the proposed framework (obtained via automatic differentiation) are also characterized  in the same plot. Specifically, these plots correspond to the computation times for the $1 \times 12$ gradient of the mean distance of all contact points with respect to the 6D SE(3) representations for the poses of the armadillos. It can be seen that the gradient computation is also fast (i.e., approximately half the speed of the original routine) and suitable for massive vectorization.

Please refer to the supplementary material for additional experiments that further characterize the proposed method.

\section{Discussion and Limitations}
In general, the proposed framework can be flexibly configured by altering any subset of its interpretable parameters (e.g. number and detail of the SDF primitives and mesh resolutions, which contact types to activate, top K numbers for vertex and edge selection, smoothing coefficients for all approximations) to better suit the needs of a down-stream control algorithm. For example, all smoothing parameters can be set very small for obtaining forward rollouts with high physical realism, and gradient computations can then use larger smoothing coefficients to prevent pathologies due to discontinuities. Similarly, the number of primitives can be reduced to trade-off geometry complexity with vectorizability, to facilitate applications such as coarse sweeps over hyper parameter grids or massively parallelized line-searches.

An important limitation of the proposed framework is that it expects a significant degree of expertise on the user side when building a simulation. For example: i)~picking the correct SDF primitives for efficient shape abstraction, ii)~making sure that there is no major discrepancy between the SDF and mesh representations of the same surface to avoid artifacts, iii)~setting the numbers $(N_1, N_2, M_1, M_2)$ for top-K selection appropriately. For (i), developing methods (similar to \cite{liu2023marching}) to automate the decomposition of arbitrary geometries into a mixture of SQ, CP, and OPC primitives in the most efficient way constitutes an interesting research direction. For (ii) and (iii), the supplementary material contains extended discussions on heuristic routines to determine $(N_1, N_2, M_1, M_2)$, as well as best practices and safeguard mechanisms to prevent any potential artifacts due to mesh-SDF discrepancies.

\section{Conclusion}
We proposed a contact manifold generation routine that is smoothly differentiable, fast, and efficiently vectorizable. This was achieved via two main technical contributions. The first is a representative set of analytical primitives to build SDFs, which are then used for two purposes: i)~as an acceleration structure to differentiably select V-S and E-E pairs close to contact via soft top-K operations, and ii)~to assign signed distances and normals to V-S and E-E contacts in a way that preserves curvature information (unlike traditional V-F and E-E contact routines between planar facets). The second is a routine to make witness point computations between V-F and E-E pairs smoothly differentiable via an analytical active-set method to solve the associated QPs, as well as differentiable schemes to assign signed distances, normals, and activity scores to them. The presented characterization experiments benchmark the proposed routine and compare it to the well-established MJX simulator. We observe a speedup by orders of magnitude. For future work, the most immediate direction is to use the proposed framework to develop efficient algorithms for optimal control and reinforcement learning that make use of first and second order information (gradients and Hessians) and exploit massive vectorizability.

\section*{Acknowledgments}
This work was supported by the ERC - 101045454 REAL-RL grant and the German Federal Ministry
of Education and Research (BMBF) through the Tübingen AI Center (FKZ: 01IS18039B). Georg
Martius is a member of the Machine Learning Cluster of Excellence, EXC number 2064/1 – Project
number 390727645. The authors thank the International Max
Planck Research School for Intelligent Systems (IMPRS-IS) for supporting Onur Beker and Anselm Paulus.


\bibliographystyle{plainnat}
\bibliography{references}

\newpage
\twocolumn[
\centering
\section*{\Large Supplementary Material}
\vspace{2em}
]

\setcounter{section}{0}

\section{Additional Experiments and Visualizations}
\begin{figure}[h]
\centering
\includegraphics[width=0.8\columnwidth]{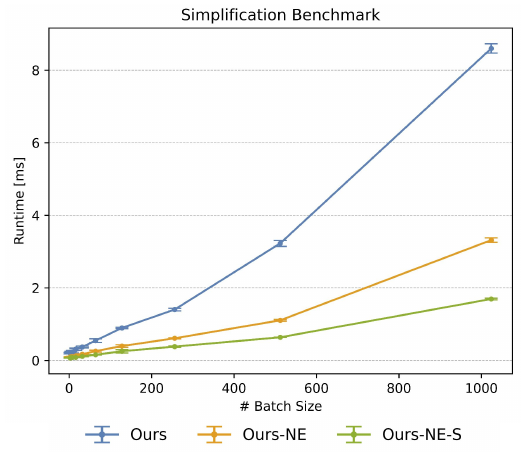}
\caption{A plot demonstrating how possible simplifications within the proposed contact manifold generation framework effect runtimes across different vectorization batch sizes. \emph{Ours} denotes the proposed pipeline without any simplifications, \emph{Ours-NE} denotes a setting where E-E contacts are turned off and only V-S contacts are considered, and \emph{Our-NE-S} denotes a setting where only one sided V-S contacts are created.}
\label{fig:simplifications}
\vspace{-6.5mm}
\end{figure}

\subsection{Possible Simplifications to Achieve Further Speed-up}

As discussed in the main paper, a major goal of formulating physics for control is identifying the right simplifications. Within a collision detection setting, one example simplification is to turn off E-E contacts and only consider V-S contacts. The curve labeled Ours-NE in \cref{fig:simplifications} shows how the runtime of such an approach scales with respect to vectorization batchsize, in the exact same experiment configuration as the vectorizability benchmark from Fig.~7 of the main paper. One can see that this simplification brings a speed-up of factor two. A further simplification can then be made by only creating one sided V-S contacts where vertices of one surface would collide with the SDF of the other, but not vice versa. In such an asymmetric setup, the surface that can be more efficiently tessellated in finer detail (e.g. with smaller surface area) would be assigned a mesh representation and the other surface would be represented with an SDF. The curve titled Ours-NE-S in \cref{fig:simplifications} again shows that this brings an additional speed-up of factor two. The supplementary videos include simulations where E-E contacts are turned off, to demonstrate that these simplifications still maintain physical plausibility when a surface is meshed isotropically with dense enough tesselation. However one still needs to exercise caution, as when such conditions do not hold (e.g. the Mesh E-E case of the box-on-box collision from Fig.5 of the main paper, where tessellation density is low and long edges are present) these simplifications can result in artifacts.


\begin{figure*}[th]
\centering
\includegraphics[width=\textwidth]{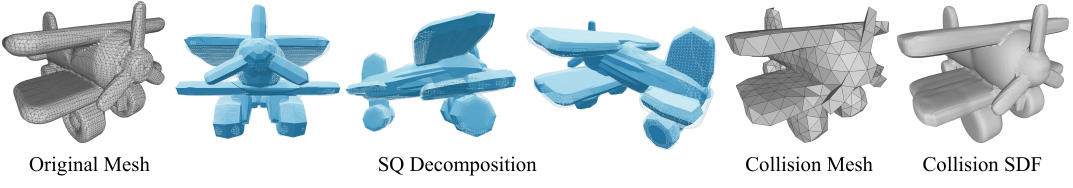}
\includegraphics[width=\textwidth]{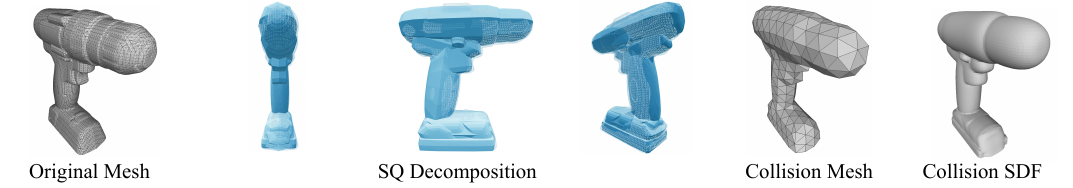}
\includegraphics[width=\textwidth]{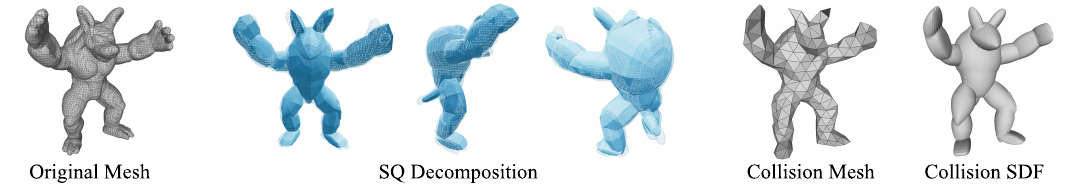}
\vspace{1mm}
\includegraphics[width=\textwidth]{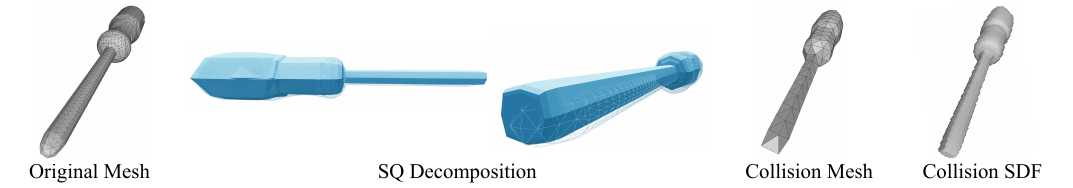}

\caption{Original high-resolution meshes, SQ primitive decompositions, as well as mesh and SDF representations (used by the proposed method) corresponding to the $4$ objects used in simulation videos. The SQ decomposition visualizations also involve an overlay of the original high-resolution mesh, to demonstrate the representation power of SQ primitives.}
\vspace{-5mm}
\label{fig:sim_vis}
\end{figure*}

\subsection{Simulation Visualizations for Different Geometries}
\cref{fig:sim_vis} shows primitive (SQ) decompositions, as well as the mesh and SDF representations for different geometries:
\begin{itemize}[leftmargin=*, label={\color{ourblue}\textbf{\textbullet}}]
    \item Toy plane from the YCB dataset \cite{calli2015ycb}. The mesh representation used for contact manifold generation contains $584$ vertices and $1746$ edges. The SDF is built from $32$ SQ primitives.
    \item Drill from the YCB dataset. The mesh representation used for contact manifold generation contains $276$ vertices and $822$ edges. The SDF is built from $17$ SQ primitives.
    \item The same armadillo mesh used in experiments from the main paper, borrowed from the Stanford 3D scanning repository \cite{krishnamurthy1996fitting}. The mesh representation used for contact manifold generation contains $305$ vertices and $605$ edges. The SDF representation is built from $18$ SQ primitives.
    \item  Screwdriver from the YCB dataset. The mesh representation used for contact manifold generation contains $165$ vertices and $489$ edges. The SDF is built from $5$ SQ primitives.
\end{itemize}
Supplementary videos in turn contain example simulations with these objects, where the contact manifold was generated using the proposed method, and the subsequent contact dynamics step was implemented using the primal cone-complementarity formulation of Mujoco. Same as the benchmarking experiments from the main paper, simulations involving these objects use no soft top-K selection for V-S contacts, and E-E contacts use a soft top-K value equal to the number of SQ primitives constituting the object (i.e. one edge per SQ). 

\subsection{Larger Versions of All Plots}
\begin{figure}[h]
\centering
\includegraphics[width=\columnwidth]{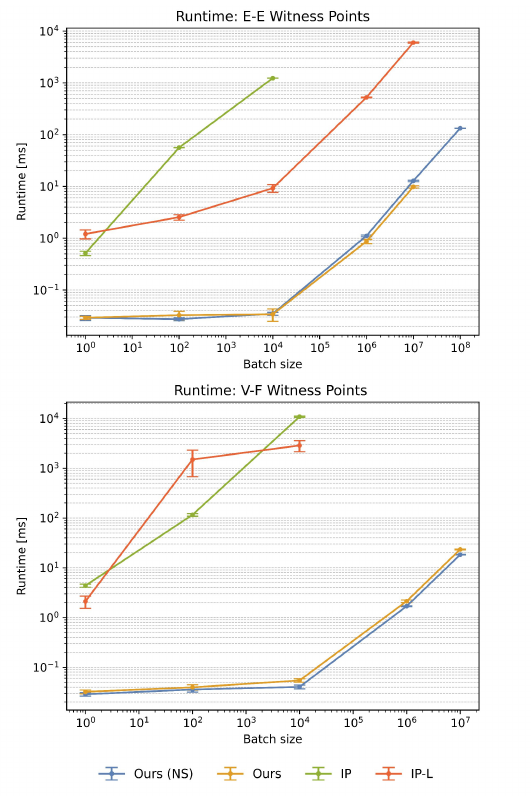}
\caption{Larger versions of plots from Fig.6 of the main paper.}
\label{fig:large_wp}
\vspace{-5mm}
\end{figure}

\begin{figure}[h]
\centering
\includegraphics[width=\columnwidth]{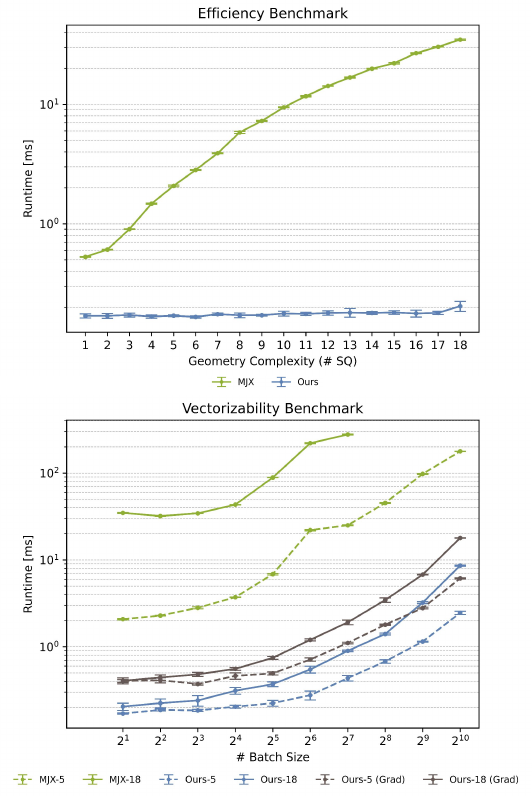}
\caption{Larger versions of plots from Fig.7 of the main paper.}
\label{fig:large_bench}
\vspace{-5mm}
\end{figure}

Please refer to \cref{fig:large_wp} and \cref{fig:large_bench} to find larger versions of plots from Fig.6 and Fig.7 of the main paper.

\section{Extended Discussions on Limitations}
\subsection{Picking the Top-K Numbers for Vertex and Edge Selection}
It is important to note that by appropriately setting the numbers $\vertexnum[\binaryidx]$ and $\edgenum[\binaryidx]$ for soft top-K selection of edges and vertices, the proposed contact manifold generation scheme can emulate both barrier methods and convex-primitive methods used in common robotics simulators. For the former barrier case, simply setting $\vertexnum[\binaryidx] = |V|$ and $\edgenum[\binaryidx] = |E|$ for both surfaces is sufficient. For the latter convex primitive case, one can compute a convex decomposition of both surfaces to obtain $K_\binaryidx$ polytopes each, and set $\vertexnum[\binaryidx]$ and $\edgenum[\binaryidx]$ to $\sim 4K_\binaryidx$ (since it is common practice to generate up to four V-F contacts or eight E-E contacts between two convex polytopes \cite{gregorius2015robust}). It is important to realize that both of these estimates for $\vertexnum[\binaryidx]$ and $\edgenum[\binaryidx]$ are very conservative upperbounds, and in most scenarios only a much smaller subset of all potential contact points are actually active (i.e. in or close to penetration). This observation is the main reason why the proposed framework uses analytical SDF primitives and soft top-K selection to implement a scheme analogous to a \say{smoothly differentiable acceleration structure} (i.e. like a BVH, kd-tree, or a spatial hash), to only select those vertices and edges that are active. While this admittedly introduces a compromise between robustness and efficiency, we argue that this is reasonable (and expected) for a formulation of physics for control. A useful heuristic process to select the numbers $\vertexnum[\binaryidx]$ and $\edgenum[\binaryidx]$ is as follows: 
\begin{itemize}[leftmargin=*, label={\color{ourblue}\textbf{\textbullet}}]
    \item Sample arbitrary SE3 configurations that create collisions between the two surfaces.
    \item Run any standard (non-differentiable) contact manifold generation process described in related work to obtain: i) the number $N$ of best-buddy pairs \cite{dekel2015best} among all E-E pairs (to determine $\edgenum[\binaryidx]$), and ii) the number $M$ of vertices that have a further penetration than all neighbors (to determine $\vertexnum[\binaryidx]$).
    \item Observe the resulting histograms for $N$ and $M$ at the end of all Monte-Carlo trials to decide on $\vertexnum[\binaryidx]$ and $\edgenum[\binaryidx]$ via a quantile that trades of efficiency and robustness (i.e. physics for control and physics for prediction).
\end{itemize}

\subsection{Safeguarding Soft Top-K Selection}
As was discussed in the background section of the main paper, soft-topK selection returns \say{soft-indices} (i.e. probabilities) $\bm\Gamma_K(\bm{x}) \in [0, 1]^{K \times D}$ that correspond to the top-K entries, and for any other arbitrary vector $\bm y \in \mathbb{R}^D$, the entries corresponding to these soft indices can be selected via a matrix product $\bm\Gamma_K(\bm{x}) \bm y \in \mathbb{R}^K$. Applying this scheme for selecting edges close to penetration involves averaging all edge endpoints (lying in rows of $\bm E_1, \bm E_2 \in \mathbb{R}^{|E| \times 3}$) using the soft-indices (i.e. $\bm\Gamma_K(\bm{x}) \bm E_1 , \bm\Gamma_K(\bm{x}) \bm E_2 \in \mathbb{R}^{K \times 3}$) This means depending on the configuration, the averaged edge endpoints can in principle leave the interiors of their respective surfaces (since the proposed method operates on non-convex surfaces). As soft-topK probabilites are often very sharp (and they get even sharper as an edge penetrates further into an opposing surface), we have not observed this to cause any artifacts. Still, it is easy and computationally cheap to include a safeguard against such a hypothetical case by scaling the activity indicators for E-E contact points using the soft-indicators $\left[ \phi_1(\edge_1^{\star}) \leq 0 \right]$ and $\left[ \phi_2(\edge_2^{\star}) \leq 0 \right]$ (which capture whether a contact point $\edge_b^{\star}$ lies within the surface described by $\phi_b$).

\subsection{Safeguarding Mesh-SDF Representation Discrepancies}
To prevent any potential artifacts in the proposed framework due to a mismatch between the SDF and mesh representations, the following practices can be adopted:
\begin{itemize}[leftmargin=*, label={\color{ourblue}\textbf{\textbullet}}]
    \item The mesh representation of the surface should be constructed such that: i) all vertices lie on the zero level set of the SDF representation, and ii) all edges should be contained inside the SDF representation (to the degree possible).
    \item After the witness points $\edge_b^{\star}=\edge_b(\alpha_b^{\star}) \text{ for } \binaryidx\in\{0,1\}$ are computed as described in subsection D-3 of the methods section, they can be projected onto the SDF surface via a fixed number of sphere tracing iterations \cite{hart1996sphere} along the direction $\nabla \phi_b(\edge_b^{\star})$. Since all of the SDF primitives introduced in subsection B of the methods section are metrically precise near the surface boundary (where witness points always lie if the mesh representation is constructed properly), the number of sphere tracing iterations needed is small enough (i.e. only $1$ for a metrically correct SDF, often $\sim\leq 5$ the for provided primitives) that the computation can be efficiently unrolled for automatic differentiation. All the subsequent computations for signed distances and normals in subsection D-3 should then use these adjusted witness points.
\end{itemize}

\section{Implementations for the Proposed V-F and E-E Witness Point Routines}
\subsection{Smooth Approximations of Non-Smooth Operators}
A subset of the smooth operators from the background section can be implemented in JAX as follows:
\begin{lstlisting}[language=Python]
def sign_s(x, eps):
  """
  x: float
  """
  return jax.nn.tanh(x/eps)

def relu_s(x, eps=0.1):
  """
  x: float
  """
  return eps * jax.nn.softplus(x / eps)

def clip_s(x, a, b, eps=0.1):
  """
  x: float
  """
  tmp1 = relu_s(x - a, eps=eps)
  tmp2 = relu_s(x - b, eps=eps)
  return a + tmp1 - tmp2

def greater_s(x, l, eps=0.1):
  """
  x: float
  """
  return jax.nn.sigmoid((x - l) / eps)

def smaller_s(x, u, eps=0.1):
  """
  x: float
  """
  return jax.nn.sigmoid((u - x) / eps)

def within_s(x, l, u, eps=0.1):
  """
  x: float
  """
  return greater_s(x, l, eps) * smaller_s(x, u, eps)

def argmin_s(x, eps=0.1):
  """
  x: (N,)
  """
  return jax.nn.softmax(-x/eps, axis=0)
\end{lstlisting}
\subsection{E-E Witness Point Routine}
The proposed smoothly differentiable active set routine to solve the E-E witness point QP (together with the utilities to then create the witness points) can be implemented as follows:
\begin{lstlisting}[language=Python]
def solve_2by2_bqp(Q, c, eps_clip=0.1, eps_min=0.1, eps_comp=0.1):
  """
  Smooth active set routine to solve a 
  2-by-2 box-constrained QP.
  
  Q: (2,2)
  c: (2,)
  """
  
    # Pre-compute necessary quantities
    Q1 = Q[0, 0]
    Q2 = Q[0, 1]
    Q3 = Q[1, 1]
    c1 = c[0]
    c2 = c[1]

    Q2_over_Q1 = Q2 / Q1
    Q2_over_Q3 = Q2 / Q3
    c1_over_Q1 = c1 / Q1
    c2_over_Q3 = c2 / Q3

    # Compute unconstrained minimum
    x1_u = (Q2 * c2_over_Q3 - c1) / (Q1 - Q2 * Q2_over_Q3)
    x2_u = (Q2 * c1_over_Q1 - c2) / (Q3 - Q2 * Q2_over_Q1)
    x_u = jp.array([x1_u, x2_u])

    # Compute all minima along the 4
    # edges of the box constraints
    x1_1_x2_u = clip_s(-(Q2_over_Q3 + c2_over_Q3), 0.0, 1.0, eps_clip)
    x1_0_x2_u = clip_s(-c2_over_Q3, 0.0, 1.0, eps_clip)
    x1_u_x2_1 = clip_s(-(Q2_over_Q1 + c1_over_Q1), 0.0, 1.0, eps_clip)
    x1_u_x2_0 = clip_s(-c1_over_Q1, 0.0, 1.0, eps_clip)

    # Compute costs for the 4 minima
    # along the box constraint edges
    x1_1_x2_u_cost = (
        0.5 * (Q1 + 2 * Q2 * x1_1_x2_u + Q3 * x1_1_x2_u * x1_1_x2_u)
        + c1
        + c2 * x1_1_x2_u
    )
    x1_0_x2_u_cost = 0.5 * Q3 * x1_0_x2_u * x1_0_x2_u + c2 * x1_0_x2_u
    x1_u_x2_1_cost = (
        0.5 * (Q1 * x1_u_x2_1 * x1_u_x2_1 + 2 * Q2 * x1_u_x2_1 + Q3)
        + c1 * x1_u_x2_1
        + c2
    )
    x1_u_x2_0_cost = 0.5 * Q1 * x1_u_x2_0 * x1_u_x2_0 + c1 * x1_0_x2_u

    # Soft-argminimum to select the smallest
    # among the constrained minima
    x1_1_x2_u_array = jp.array([1.0, x1_1_x2_u])
    x1_0_x2_u_array = jp.array([0.0, x1_0_x2_u])
    x1_u_x2_1_array = jp.array([x1_u_x2_1, 1.0])
    x1_u_x2_0_array = jp.array([x1_u_x2_0, 0.0])

    costs = jp.array([x1_1_x2_u_cost, x1_0_x2_u_cost, x1_u_x2_1_cost, x1_u_x2_0_cost])
    arrays = jp.array(
        [x1_1_x2_u_array, x1_0_x2_u_array, x1_u_x2_1_array, x1_u_x2_0_array]
    )
    min_probs = argmin_s(costs, eps_min)
    x_c = (arrays * min_probs[:, None]).sum(axis=0)

    # Compute the soft indicator for whether 
    # constraints are active
    constraint_inactive = within_s(x1_u, 0, 1, eps_comp) * within_s(
        x2_u, 0, 1, eps_comp
    )
    constraint_active = 1 - constraint_inactive

    # Pick whether the unconstrained or constrained
    # minimum will be returned
    x_opt = x_u * constraint_inactive + x_c * constraint_active

    return x_opt, constraint_inactive

def interpolate_edge(u, p):
    """
    u: float in [0, 1]
    p: (2, 3)
    """

    return p[0] + (p[1] - p[0]) * u
    
def wp_E_E_bqp(
    e1_p,
    e2_p,
    w_reg=0.01,
    eps_clip=0.1,
    eps_min=0.1,
    eps_comp=0.1,
):

  """
  Computes the two witness points of an E-E pair.
  
  e1_p: (2,3), edge corner-points
  e2_p: (2,3), edge corner-points
  """
    A = jp.stack([e1_p[1] - e1_p[0], e2_p[0] - e2_p[1]], axis=1)
    b = e1_p[0] - e2_p[0]

    Q = A.T @ A + w_reg * jp.eye(2)
    c = b.T @ A - w_reg * jp.array([0.5, 0.5])

    u_u, gamma_con = solve_2by2_bqp_s(
        Q,
        c,
        eps_clip=eps_clip,
        eps_min=eps_min,
        eps_comp=eps_comp,
    )

    witness_p1 = interpolate_edge(u_u[0], e1_p)
    witness_p2 = interpolate_edge(u_u[1], e2_p)

    return witness_p1, witness_p2, u_u, gamma_con
\end{lstlisting}

\subsection{V-F Witness Point Routine}
Analogous routines to the previous section can be implemented for V-F contacts as follows:
\begin{lstlisting}[language=Python]
def V_E_proj(v, e_p):
    e_t = normalize(e_p[1] - e_p[0])
    return clip_s((v - e_p[0]).T @ e_t, 0, 1) * e_t + e_p[0]

def V_T_proj_s(v, T_p, eps_clip=0.1, eps_min=0.1, eps_comp=0.1):

    # Pre-compute necessary quantities
    delta_10 = T_p[1] - T_p[0]
    delta_21 = T_p[2] - T_p[1]
    delta_20 = T_p[2] - T_p[0]
    delta_v0 = v - T_p[0]
    delta_v1 = v - T_p[1]
    e_t1 = normalize(delta_10)
    e_t2 = normalize(delta_21)
    e_t3 = normalize(delta_20)

    # Compute all 3 minima along the constraints
    # by projecting the vertex onto triangle edges
    vp_e1 = clip_s((delta_v0 * e_t1).sum(), 0, 1, eps_clip) * e_t1 + T_p[0]
    vp_e2 = clip_s((delta_v1 * e_t2).sum(), 0, 1, eps_clip) * e_t2 + T_p[1]
    vp_e3 = clip_s((delta_v0 * e_t3).sum(), 0, 1, eps_clip) * e_t3 + T_p[0]

    # Soft-argminimum to select the smallest
    # among the constrained minima
    costs = jp.array(
        [norm(v - vp_e1), norm(v - vp_e2), norm(v - vp_e3)]
    )
    arrays = jp.array([vp_e1, vp_e2, vp_e3])
    min_probs = argmin_s(costs, eps_min)
    vp_c = (arrays * min_probs[:, None]).sum(axis=0)

    # Compute the unconstrained minimum by
    # projecting the vertex onto the triangle plane
    n_bar = jp.cross(delta_10, delta_20)
    n, n_bar_norm = normalize_with_norm(n_bar)
    delta_vp_0 = delta_v0 - (delta_v0 * n).sum() * n
    vp_u = delta_vp_0 + T_p[0]

    # Check whether the unconstrained minimum is within
    # the constraints, using barycentric coordinates
    v = (jp.cross(delta_10, delta_vp_0) * n).sum() / n_bar_norm
    u = (jp.cross(delta_vp_0, delta_20) * n).sum() / n_bar_norm
    w = 1 - u - v

    constraint_inactive = (
        within_s(u, 0, 1, eps_comp)
        * within_s(v, 0, 1, eps_comp)
        * within_s(w, 0, 1, eps_comp)
    )
    constraint_active = 1 - constraint_inactive

    # Pick whether the unconstrained or constrained
    # minimum will be returned
    vp = vp_u * constraint_inactive + vp_c * constraint_active

    return vp
    
\end{lstlisting}


\begin{table*}[t]
\centering
\small
\setlength{\tabcolsep}{6pt}
\renewcommand{\arraystretch}{1.15}
\begin{tabular}{@{}llp{0.56\linewidth}@{}}
\toprule
\textbf{ID} & \textbf{File} & \textbf{Description} \\
\midrule

\multicolumn{3}{@{}l@{}}{\textbf{Folder:} \texttt{armadillo\_armadillo\_rollout/}} \\
\midrule
A1 & \texttt{angle\_1.mp4} & Armadillo--Armadillo collision rollout (view angle 1). Shows contact manifold behavior on a non-convex mesh pair. \\
A2 & \texttt{angle\_2.mp4} & Same rollout as A1 from view angle 2. \\
A3 & \texttt{angle\_3.mp4} & Same rollout as A1 from view angle 3. \\
A4 & \texttt{angle\_4.mp4} & Same rollout as A1 from view angle 4. \\
A5 & \texttt{angle\_1\_E\_E\_contacts\_off.mp4} & Same setup as A1, but with E--E contacts disabled (V--S only). Demonstrates the ``Ours-NE'' simplification discussed in the supplementary material \cref{fig:simplifications}. \\
A6 & \texttt{E\_E\_on\_off\_comparison.mp4} & Side-by-side comparison with E--E contacts enabled vs.\ disabled for Armadillo--Armadillo. \\
A7 & \texttt{interactive\_visualizer\_demo.mp4} & Demo of the interactive visualizer used to inspect contact manifolds, normals, and activity indicators during simulation. \\

\midrule
\multicolumn{3}{@{}l@{}}{\textbf{Folder:} \texttt{plane\_drill\_rollout/}} \\
\midrule
P1 & \texttt{angle\_1.mp4} & Plane (YCB) -- Drill (YCB) collision rollout (view angle 1). \\
P2 & \texttt{angle\_2.mp4} & Same rollout as P1 from view angle 2. \\
P3 & \texttt{angle\_3.mp4} & Same rollout as P1 from view angle 3. \\
P4 & \texttt{angle\_4.mp4} & Same rollout as P1 from view angle 4. \\
P5 & \texttt{angle\_1\_E\_E\_contacts\_off.mp4} & Same setup as P1, but with E--E contacts disabled (V--S only). \\
P6 & \texttt{E\_E\_on\_off\_comparison.mp4} & Side-by-side comparison with E--E contacts enabled vs.\ disabled for Plane--Drill. \\
P7 & \texttt{interactive\_visualizer\_demo.mp4} & Interactive visualizer demo for Plane--Drill contact inspection. \\

\midrule
\multicolumn{3}{@{}l@{}}{\textbf{Folder:} \texttt{drill\_screwdriver\_rollout/}} \\
\midrule
D1 & \texttt{angle\_1.mp4} & Drill (YCB) -- Screwdriver (YCB) collision rollout (view angle 1). \\
D2 & \texttt{angle\_2.mp4} & Same rollout as D1 from view angle 2. \\
D3 & \texttt{angle\_3.mp4} & Same rollout as D1 from view angle 3. \\
D4 & \texttt{angle\_4.mp4} & Same rollout as D1 from view angle 4. \\
D5 & \texttt{angle\_1\_E\_E\_contacts\_off.mp4} & Same setup as D1, but with E--E contacts disabled (V--S only). \\
D6 & \texttt{E\_E\_on\_off\_comparison.mp4} & Side-by-side comparison with E--E contacts enabled vs.\ disabled for Drill--Screwdriver. \\
D7 & \texttt{interactive\_visualizer\_demo.mp4} & Interactive visualizer demo for Drill--Screwdriver contact inspection. \\

\bottomrule
\end{tabular}
\caption{Supplementary videos. Each scenario is shown from four camera angles, plus an ablation with E--E contacts disabled (V--S only), an E--E on/off comparison, and an interactive visualizer demonstration.}
\label{tab:supplementary_videos}
\end{table*}

\end{document}